\documentclass{article}

\usepackage{amsmath}
\usepackage{booktabs}
\usepackage{cite} 
\usepackage[breaklinks=true]{hyperref}
\usepackage[utf8]{inputenc}
\usepackage{multirow}
\usepackage{titling}
\usepackage{url}
\usepackage[english]{babel}
\usepackage{breakurl}
\usepackage{underscore}
\usepackage{fullpage}
\usepackage{listings}
\usepackage{parskip}
\usepackage{authblk}
\usepackage{float}
\usepackage[title]{appendix}
\usepackage{etoolbox}
\usepackage{xcolor}

\patchcmd{\appendices}{\quad}{: }{}{}

\newcommand{\subtitle}[1]{%
  \posttitle{%
    \par\end{center}
    \begin{center}\large#1\end{center}
    \vskip0.5em}%
}

\newcommand{\email}[1]{\href{mailto:#1}{\nolinkurl{#1}}}

\hypersetup{colorlinks=true}

\title{ZR-2021VG: Zero-Resource Speech Challenge, Visually-Grounded Language Modelling track, 2021 edition}
\subtitle{Version 2.0 -- final for NeurIPS}
\date{July 13, 2021}
\author[a]{Afra Alishahi}
\author[a]{Grzegorz Chrupała}
\author[b]{Alejandrina Cristia}
\author[b]{Emmanuel Dupoux}
\author[a]{Bertrand Higy}
\author[b]{Marvin Lavechin}
\author[c,d]{Okko Räsänen}
\author[e]{Chen Yu}
\affil[a]{Dept. of Cognitive Science and AI, Tilburg University, Netherlands}
\affil[b]{Laboratoire de Sciences Cognitives et Psycholinguistique, ENS, Paris, France}
\affil[c]{Unit of Computing Sciences, Tampere University, Finland}
\affil[d]{Dept. Signal Processing and Acoustics, Aalto University, Finland}
\affil[e]{Dept. of Psychology, University of Texas at Austin}

\begin{document}
\maketitle

\section{Introduction}
This document introduces the visually-grounded spoken language modeling track of the ZeroSpeech 2021 challenge. In this track, participants are asked to use audiovisual materials during training. Evaluation is identical to the speech-only track at ZeroSpeech 2021. 

 Learning to comprehend and produce spoken languages is one of the hallmarks of human cognition, and the importance of speech communication also makes speech-based capabilities central to AI development. Modern automatic speech recognition (ASR) systems largely rely on supervised training, where input speech is paired with corresponding phonetic annotations or text transcripts. While this approach has produced great results in high-resource languages such as English, deployment of similar systems for low-resource environments such as small language communities or even unwritten languages is difficult. In addition, a mismatch between conversational speech and large-scale text data still exists even in high-resource languages. 
 
 In contrast to ASR, human children achieve their language skills without direct supervision or detailed feedback, simply interacting with their physical and linguistic environments. Moreover, these experiences are essentially \textit{multimodal}: children not only hear speech of their caregivers, but concurrently observe the world through a number of senses. Instead of perceiving speech and the corresponding text, they learn from speech in the context of different everyday multimodal communicative scenarios. 
 Compared with audio data only, audiovisual data contain more statistical regularities at multiple levels: there are regularities within audio and visual data respectively; and there are also cross-modal regularities at both the acoustic and semantic levels. For example, the presence of a dog in a visual scene is likely to co-occur with both barking sounds at the acoustic level and the spoken word “dog” (and other words semantically related to dog) at the semantic level. The presence of a dog can be used as a supervisory signal to train language learning systems to parse and segment words that consistently occur in the same context, and semantically group words that occur in similar visual contexts. 
 In order to develop AI systems or models of human learning with similar multimodal language learning skills, a number of models and learning algorithms have been proposed throughout the years (e.g., \cite{roy2002learning,Yu_Ballard_2004,tenbosch_2009,Driesen_2011,Rasanen_Rasilo_2015, Mangin_2015}). These systems have used various types of speech data with simulated or robot vision-based visual input. 
 
 However, only the recent advances in deep learning have scaled up the capabilities of audiovisual systems to a level where they can start to capture the relationships between realistic visual data (e.g., photographs or videos) and language related to the visual scene. These models were first developed for captions descriptive of images \cite{socher2014grounded,karpathy2015deep} and more recently for spoken descriptions of the images (e.g., \cite{harwath2015deep,harwath2016unsupervised,synnaeve2014learning,chrupala2017representations,kamper2019semantic}). In this context, it is of great interest how the representations emerging from training of such multimodal models relate to the known linguistic structure of the input language (e.g., \cite{chrupala2017representations,alishahi2017encoding,harwath2019learning,havard2019models,havard2019word}), and how such methods can support (or replace) purely audio-based representation learning approaches (e.g., \cite{Chung2019a,oord2018cpc}). In other words, it would be highly useful if unsupervised learning from multimodal data could be used to acquire language representations such as phone(me)s and words without access to transcribed training data in the given language---units that can be then used as a basis for many other language processing tasks. However, the research in this direction is still young and largely driven by a few research groups \cite{chrupala2021visually}. In addition, there are no standardized evaluation metrics or a common benchmark to compare different methodological approaches and thereby to drive the research  in this area forward. 
 
 The goal of the ZR-2021VG track is to tackle the issue of multimodal language learning.  In contrast to the earlier Zerospeech-challenges \cite{versteegh2015zero,dunbar2017zero,dunbar2019zero,nguyen2020zero} that have purely focused on audio-based learning of linguistic representations (including the speech-based track of the current challenge, on which we build), ZR-2021VG takes a step towards multimodal language learning by asking participants to train on audiovisual data. The aim of the challenge is to learn phonemic, lexical, syntactic, and semantic representations of speech with the help of supporting visual information, as evaluated by standardized evaluation protocols. As a result, the challenge aims to bring together researchers from speech technology, natural language processing, computer vision, and machine learning  to work on multimodal language learning, thereby advancing the state-of-the-art in audiovisual learning algorithms and providing new knowledge on how visual information may support unsupervised learning of linguistic patterns from speech.
 


\section{Registration}
Participation is open and free. Registration is done via the ZR2021 website \url{https://zerospeech.com/2021/index.html#zrc2021}. The Challenge will be submitted as such to several conferences, and participants are encouraged to submit a paper to them. The first conference is NeurIPS; more information on participation is available from \url{https://zerospeech.com/2021/index.html#neurips-competition}

For any issues or questions, we recommend to registrants that they join our mailing list:
\begin{enumerate}
    \item Make sure you're logged into google (you may need to get your email account registered with them here: \url{https://www.google.com/account/about})
    \item Go to \url{https://groups.google.com/g/zrvg}
    \item Ask to join.
\end{enumerate}

A moderator will approve your request within a week.

\section{Task}
\subsection{Task definition}

The challenge is to learn spoken language representations from raw audio-visual input, without any annotation or written transcriptions. Systems are allowed to use the raw audio and image/video (or features extracted from them, such as MFCCs for the audio) of the training set(s) as input. The goal is to use the raw multimodal input data to discover discrete linguistic units at phonetic and word levels, which will be evaluated via a set of black-box, zero-shot metrics probing for the quality of the learned models at different linguistic levels, including phonetics, lexicon, syntax and semantics. See Section \ref{task:eval} for more information.

Any approach that exploits the correspondence between the two modalities without relying on supervision based on linguistic annotations of the speech or visual signal is allowed (see Section \ref{sec:rules} for details). A range of objectives could be used for this purpose, including (but not limited to):
\begin{itemize}
\item  generating the visual representations from the audio or vice versa;
\item  predicting the next audio frame  with the visual input as an additional signal; 
\item  training language models on pseudo-text that has been generated with the help of the visual input; 
\item  inducing linguistic and visual representations that are geometrically similar in a shared semantic space. 
\end{itemize}

In order to take into account the computing resources of participants, we distinguish categories of submissions in two tracks based on the type of model and resources employed (see Section \ref{task:model}). 
Additionally, we may highlight submissions that explore innovative architectures, successfully use particularly small amounts of training data, or have other unusual features which contribute to fostering new ideas and gaining collective insight into this interesting problem. 

\subsection{Model conditions}\label{task:model} 

There are no restrictions or conditions on the models: if they are unsupervised, they are eligible.

In order to take into account the computing resources of participants, we ask participants to report their models' resource budget. As in ZR2021, you calculate your budget as number of hours $\times$ number of GPUs. Additionally, if relevant, report Akaike’s Information Criterion and Bayesian Information Criterion; and/or number of parameters and number of pre-trained parameters. (If you don't know what that means, it probably is not relevant.)

We also ask participants to report how much data (in number of speech hours, number of images, hours of video) they used for training their submitted models.

One overall restriction on models comes from the evaluation, which we want to align with the audio-only track of the Zerospeech-2021 challenge. The evaluation procedure focuses on linguistic representations (ignoring the visual representation), and will do so using \textbf{unimodal} audio data. This means that the models submitted to the challenge should be able to process audio data independently of visual data.

\subsection{Evaluation conditions}\label{task:eval}

We will now briefly describe the metrics and refer the reader to \cite{nguyen2020zero} for more details.

Each metric evaluates models at a different linguistic level:
\begin{itemize}
    \item \textbf{Phonetic (ABX metric)}. The ABX metric allows to assess the degree of discriminability between two categories $A$ and $B$ given a representation, where in our case $A$ and $B$ correspond to minimal pairs of triphones such as /beg/ and /bag/ (differing only in their middle phoneme). Given a triple of stimuli $(a, b, x)$ where $a$ and $x$ belong to the same category $A$ and $b$ to a different category $B$, and given a distance $d(\cdot, \cdot)$ between pairs of stimuli, discrimination is considered successful if $d(a, x) < d(b, x)$. The ABX score is obtained by aggregating scores across all minimal triphone pairs. 
    \item \textbf{Lexical (spot-the-word metric)}. This metric evaluates the possibility of discriminating between real words and nonwords by associating a probability to each of them (with the expectation that a higher probability is associated with the real word). The score is computed as the average discrimination accuracy across all pairs of words and nonwords. 
    \item \textbf{Syntactic (acceptability metric)}. Similarly to the spot-the-word metric, the acceptability metric computes a discrimination accuracy between grammatical and ungrammatical categories, based on a probability associated to each of them. The pairs of sentences we use are representative of 68 different syntactic paradigms belonging to twelve broad categories. 
    \item \textbf{Semantic (similarity metric)}. This metric compares the similarity between the representations of pairs of words to human similarity judgements. The metric is defined as the Spearman's rank correlation coefficient between the two similarity scores. 
\end{itemize}

\section{Baseline models}\label{sec:benchmarks}

To represent a variety of participants, two baseline models have been produced.\footnote{Code and detailed instructions to repoduce our results can be found at \url{https://github.com/bhigy/zr-2021vg\_baseline}} One uses a low budget (72 GPU hours); the other uses a high budget (165 GPU hours), corresponding to the following submission tracks:
\begin{itemize}
    \item Track A: Low budget models. Low budget submissions use smaller models which can be trained with lower GPU memory and training hours (about 100 GPU hours maximum).
    \item Track B: High budget models. High budget submissions use more GPU memory and/or more training time.
\end{itemize}

Our baselines are directly inspired by the baselines used in the audio-only track. The main difference is that we incorporate a visually grounded (VG) model to learn our speech representations. Those representations are then fed to the language model through K-means clustering. The low-budget baseline completely replaces the contrastive predictive model (CPC) with the VG model. The high-budget baseline, on the other hand, adds the VG model on top of the CPC model.

\subsection{Data}

The baselines are trained with two datasets:
\begin{itemize}
    \item SpokenCOCO \cite{hsu_text-free_2020} for the VG models.
    \item LibriSpeech \cite{Panayotov2015librispeech} for K-means clustering (100h from the \emph{train-clean-100} subset) and the language models (960h from the \emph{train-clean-100}, \emph{train-clean-360} and \emph{train-other-500} subsets).
\end{itemize}

Image features used to train the VG model are extracted from the preclassification layer of a frozen ResNet-152 model \cite{he_deep_2016} pretrained on ImageNet \cite{deng_imagenet_2009}. We follow \cite{merkx_language_2019} and use the mean feature vector over ten crops of each image.

The acoustic feature vectors used by the low-budget baseline are composed of 12 mel-frequency cepstral coefficients (MFCCs) and log energy, with first and second derivatives, resulting in 39-dimensional vectors. They are computed over windows of 25 ms of speech, with 10 ms shift. The high-budget baseline uses features extracted from a pretrained CPC model that works on raw waveform.

\subsection{Architecture}

\subsubsection{Low-budget baseline}

The VG model follows the speech-image architecture described in \cite{chrupala_symbolic_2019, higy_textual_2020}. It is composed of visual and speech encoders which each extract fixed length embeddings from the visual and audio input respectively.

The image encoder is composed of a single linear layer projecting the image features into a shared semantic embedding space (dimension 2048), followed by a normalization layer ($\ell^2$ norm).

The speech encoder is composed of a 1D convolutional layer (kernel size 6, stride 2 and 64 output channels), followed by bidirectional gated recurrent units (GRUs) \cite{cho_learning_2014} (4 layers, hidden
state of dimension 1024). A vectorial attention layer is then used to
convert the variable length input sequence to a fixed-size vector of dimension
2048. Finally, a normalization layer ($\ell^2$ norm) is applied.

Once the model is trained, we extract activations of the $1^{\text{st}}$ recurrent layer of the speech encoder and cluster them with K-means (50 clusters). The quantized activations are then used to train a language model corresponding to the BERT-small architecture from the audio-only track.

\subsubsection{High-budget baseline}

The high-budget baseline is essentally the same as the low-budget baseline with following exceptions:
\begin{itemize}
    \item The VG model is trained on features extracted from the pretained CPC-small model from the audio-only track. The activations of the last recurrent layer of the CPC model are used.
    \item The language model corresponds to the BERT-big architecture from the audio-only track.
\end{itemize}

\subsection{Evaluation}

We evaluated the models on the four metrics introduced in \cite{nguyen2020zero}. The phonetic scores are computed with cosine distance on continuous representation (i.e. before quantization) extracted from the $1^{\text{st}}$ (resp. $2^\text{{nd}}$) recurrent layer of the VG (resp. CPC) model for the low-budget (resp. high-budget) baseline. Lexical and syntactic metrics rely on pseudo-probabilities obtained from the language model of each pipeline. Finally, the semantic scores are computed using activations extracted from the last (resp. $4^{\text{th}}$) recurrent layer of the language model for the low-budget (resp. high-budget) audio-only baseline, while the output of the attention layer was used for both visually-grounded models. All semantic scores are obtained with max pooling\footnote{For the visually-grounded models, the choice of the pooling mechanism doesn't affect the results as the output of the attention layer is a fixed length vector and not a sequence.} and the cosine distance to evaluate the similarity between activations. 

Table \ref{tab:results} summarizes the results obtained with the two visually-grounded baselines as well as the two baselines from the audio-only track.

While the audio-only baselines tend to perform better than their visually-grounded counterparts on phonetic, lexical and syntactic metrics, the latter obtain higher semantic scores overall. One exception is the un-weighted semantic similarity scores on the test set where both visually-grounded baselines get very low scores. Upon inspection, this can be attributed to unstability in the scores obtained on the smallest sub-datasets of the test set. This motivated the introduction of a weighted version of the metric (where each sub-score is weighted by the size of the dataset it is obtained from) which shows much more stability.

We can also observe that the high-budget visually-grounded baseline, by using CPC-small as a feature extractor, largely closes the gap with the high-budget audio-only baseline on the phonetic, lexical and syntactic metrics (especially the last two).

\setlength{\tabcolsep}{3pt}   
\begin{table}
    \centering
    \small
    \begin{tabular}{ccccccccccccc}
    	\toprule
        \multirow{3}{*}{Track} & \multirow{3}{*}{Budget} &\multirow{3}{*}{Set} & \multicolumn{4}{c}{Phonetic} & \multirow{3}{*}{Lexical} & \multirow{3}{*}{Syntactic} & \multicolumn{4}{c}{Semantic} \\
        \cline{4-7} \cline{10-13}
        & & & \multicolumn{2}{c}{Within} & \multicolumn{2}{c}{Across} & & & \multicolumn{2}{c}{Un-weighted} & \multicolumn{2}{c}{Weighted} \\
        \cline{4-5} \cline{6-7} \cline{10-11} \cline{12-13}
		& & & clean & other & clean & other & & & synth. & libri. & synth. & libri. \\
		\midrule
        \multirow{4}{*}{Audio-only} & \multirow{2}{*}{Low} & dev & \textbf{0.03} & \textbf{0.05} & \textbf{0.04} & \textbf{0.08} & 0.61 & 0.52 & 4.42 & 7.07 & 4.42 & 7.07 \\
        & & test & 0.03 & 0.05 & 0.04 & 0.08 & 0.61 & 0.53 & 7.35 & 2.38 & 7.31 & 5.82 \\
		& \multirow{2}{*}{High} & dev & \textbf{0.03} & \textbf{0.05} & \textbf{0.04} & \textbf{0.08} & \textbf{0.68} & \textbf{0.56} & 6.25 & 4.35 & 6.25 & 4.35 \\
        & & test & 0.03 & 0.05 & 0.04 & 0.08 & 0.68 & 0.56 & 5.17 & 2.48 & 3.19 & 1.32 \\
        \midrule
        \multirow{4}{*}{Visually-grounded} & \multirow{2}{*}{Low}  & dev  & 0.09 & 0.10 & 0.11 & 0.15 & 0.53 & 0.53 & \textbf{9.65} & 12.61 & \textbf{9.65}  & 12.61 \\
        & & test & 0.08 & 0.11 & 0.11 & 0.15 & 0.53 & 0.53 & 9.71 & 0.16 & 12.57  & 12.55 \\
		& \multirow{2}{*}{High} & dev  & 0.06 & 0.07 & 0.07 & 0.11 & 0.67 & 0.55 & 9.60 & \textbf{15.09} & 9.60  & \textbf{15.09} \\
        & & test & 0.05 & 0.07 & 0.07 & 0.12 & 0.67 & 0.55 & 9.99 & -0.10 & 13.46 & 10.96 \\
        \bottomrule
    \end{tabular}
    \caption{Overall performance of the audio-only and the visually grounded baselines on dev and test sets. Phonetic scores are reported in terms of within and across speakers ABX error rates (lower is better). Lexical and syntactic scores represent accuracies computed on the pseudo-probability of spotting the right stimuli (higher is better). The semantic similarity score is reported as the Spearman's rank correlation coefficient between the distance scores returned by the model and the true human scores (higher is better) Numbers in bold indicate best scores on the development set for each metric.}
    \label{tab:results}
\end{table}

\section{Data}
\label{sec:data}
\subsection{Training data}
\label{sec:train}

Training data are not provided. Instead, ZR-VG2021 participants may use any publicly available, private, or proprietary data (audio or visual stream or snapshots; instructional videos with voiceovers; see Appendix \ref{app:sources} for examples) to train their systems provided. However, synthetic speech that is generated from text is not allowed.
We strongly encourage the use of public data, as this improves interpretability and cumulativity of results.  In addition, the following corpora cannot be used because they are part of the evaluation set (for more details, see ZR2021's challenge description \cite{nguyen2020zero}):
\begin{itemize}
    \item sWUGGY (from ZR2021)
    \item sSIMI (from ZR2021)
    \item sBLIMP (from ZR2021)
\end{itemize}

A good starting point for people new to working with audio-visual data is SpokenCOCO dataset, which was also used to train our baselines.

Participants are responsible for documenting precisely the training data they used in any final submissions, as part of their Methods section, including any pre-processing (such as correcting or adding voice activity detection or denoising. We also ask that participants overtly describe their training data in terms of how similar to natural human visual and audio flow they are. For instance, Librispeech is read speech and thus more formal and simpler than natural audio flow, whereas the VanDam corpora \cite{vandam_homebank_2015, vandam_homebank_2015-1} was captured with a wearable, thus representing natural audio conditions from a first-person perspective.

\subsection{Development and test data}
Scoring will be done by the ZR team, as part of their general yearly challenge. Dev/test data are therefore not defined here. Please see \cite{nguyen2020zero} for information. 

\section{Evaluation rules}
\label{sec:rules}
The ZR-VG2021 is an open evaluation challenge. Participants must agree to work respecting the following rules:

\begin{itemize}
    \item You should describe training and system data thoroughly, as well as the computational resources used, for your final systems in any submission. Please see the Appendix \ref{app:checklist} for the full list.
    
    \item You should do your best to share training data and code publicly, for example by depositing them in a scientific archive such as the Open Science Framework, osf.io, which contains a Github plugin.
    
    \item You should not use the prohibited corpora listed in Section \ref{sec:train}.
    
    \item You should not perform manual/human investigation of the evaluation data such as listening, manual segmentation, transcription, or other form of human annotation. Examples of such prohibited practice are speech recognition with a language model, phone(me) recognition with a supervised classifier, or using written captions or synthesized-speech captions. 
    \item You can use any automatically derived information as long as that system was not trained with linguistic labels, for example speaker diarization and denoising, or speaker or language identification.  You can also use spoken captions of pictures. 
    
    \item Training must be done in an unsupervised fashion \textit{within the linguistic domain}; i.e., no linguistic labels allowed, including those generated via ASR systems (since these have themselves been trained using linguistic labels). 
    
    \item One exception to the general rule of no supervision is for visual features that may be pre-trained with object labels (such as the ImageNet labels), but we categorically forbid the use of full captions, and we caution against using this exception to allow linguistic supervision to seep in. If you are uncertain, please contact the organizers.
\end{itemize}

Failure to abide by the rules above can lead to disqualification from the challenge, removal of existing submissions, and public remonstrations.

Submissions will be ranked per track (see Section \ref{sec:benchmarks}) on each of the four metrics described in Section \ref{task:eval} separately.

\newpage
\begin{appendices}
\section{Ideas for training data}
\label{app:sources}

As in ZR-2021, the following are training options for clean audio:

\begin{itemize}
\item \href{http://www.openslr.org/12/}{LibriSpeech} (the standard subsets: clean-100, clean-360, other-500)
\item \href{https://github.com/facebookresearch/libri-light}{Libri-light}: small (600h), med(6kh) or large(60kh).
\end{itemize}

In addition, we suggest the following audio flow corpora:

\begin{itemize}
    \item From Homebank, child-centered audio flow captured with a wearable
        \begin{itemize}
        \item Fausey, C. M., \& Mendoza, J. K. (2018). FauseyTrio-Public HomeBank Corpus. doi:10.21415/T56D7Q \url{https://homebank.talkbank.org/access/Public/FauseyTrio-Public.html}
        \item  VanDam, Mark (2018). VanDam Public 5-minute HomeBank Corpus. doi:10.21415/T5388S \url{https://homebank.talkbank.org/access/Public/VanDam-5minute.html}
        \item VanDam, Mark (2018). VanDam Public Daylong HomeBank Corpus. doi:10.21415/T5QH5N  \url{https://homebank.talkbank.org/access/Public/VanDam-Daylong.html}
        \end{itemize}
\end{itemize}

The following are some ideas for training options for clean visual data:

Here are training options for audio-video flow:

\begin{itemize}
\item \url{https://childes.talkbank.org/}
\item \href{http://youcook2.eecs.umich.edu/}{YouCook II}, 2K videos, 176 hours
\cite{zhou2018towards}
\item \href{https://srvk.github.io/how2-dataset/}{How2}, 80k instructional videos (about 2000h) with associated English subtitles and summaries that participants should NOT use. \cite{sanabria18how2}
\item \href{https://www.di.ens.fr/willow/research/howto100m/}{HowTo100M}, 136M video clips with captions sourced from 1.2M Youtube instructional videos (15 years of videos) covering 23k activies: cooking, hand crafting, personal care, gardening \cite{miech19howto100m}
\item \href{https://research.google.com/ava/}{AVA Spoken Activity Datasets} (densely labeled for {NO_SPEECH, CLEAN_SPEECH, SPEECH_WITH_MUSIC, SPEECH_WITH_NOISE}). Extracted from 192 movies publicly available on YouTube (multiple languages); 45 hours total \cite{AVA}
\end{itemize}

Audio-image flow:

\begin{itemize} 
    \item \href{https://groups.csail.mit.edu/sls/downloads/placesaudio}{MIT Places 205}: 400,000 spoken audio caption  each of which describes a different Places image;  collected from over 2,500 different speakers and covering a 40,000 word vocabulary. The average caption duration is approximately 10 seconds. Used in \cite{harwath2019learning}.
    \item \href{https://groups.csail.mit.edu/sls/downloads/flickraudio}{Flickr8K}: contains 40,000 read audio captions describing 8,000 images from Flickr. Used in \cite{harwath2015deep}.
    \item \href{https://groups.csail.mit.edu/sls/downloads/placesaudio}{SpokenCOCO}: contains approximately 600,000 read audio captions describing images from MSCOCO dataset. Used in \cite{hsu_text-free_2020}. 
\end{itemize}


\section{Q\&A}
\label{app:qa}

Q: Is it possible to use instructional videos with voiceovers?

A: Yes!

Q: I'm not interested in creating a model for the lower levels (phonetics), I only care about word learning.

A: You are welcome to use other unsupervised models at the beginning of your pipeline, for instance by taking our baseline, and replacing the late components with your own ideas for the "semantic" level sections.

Q: Can I use other data, beyond audio and visual?

A: Provided it's not text labels, you probably can. For instance, a model that is trained with 1. an image paired with 2. an audio file and 3. time-course of human gazes on the image would be acceptable

\section{Checklist for submission}
\label{app:checklist}

Have you:

\begin{itemize}
    \item played by the rules, as in Section \ref{sec:rules}?
    \item clearly stated in your paper what your budget was?
    \item reported on all 4 evaluation conditions, at least in an online appendix to your paper?
    \item described the data you used for training (in terms of sources as well as in terms of quantity)?
    \item described at least one of your models in terms of complexity and computing resources?
\end{itemize}

\bibliographystyle{IEEEtran}
\bibliography{refs}

\begin{thebibliography}{10}
\providecommand{\url}[1]{#1}
\csname url@samestyle\endcsname
\providecommand{\newblock}{\relax}
\providecommand{\bibinfo}[2]{#2}
\providecommand{\BIBentrySTDinterwordspacing}{\spaceskip=0pt\relax}
\providecommand{\BIBentryALTinterwordstretchfactor}{4}
\providecommand{\BIBentryALTinterwordspacing}{\spaceskip=\fontdimen2\font plus
\BIBentryALTinterwordstretchfactor\fontdimen3\font minus
  \fontdimen4\font\relax}
\providecommand{\BIBforeignlanguage}[2]{{%
\expandafter\ifx\csname l@#1\endcsname\relax
\typeout{** WARNING: IEEEtran.bst: No hyphenation pattern has been}%
\typeout{** loaded for the language `#1'. Using the pattern for}%
\typeout{** the default language instead.}%
\else
\language=\csname l@#1\endcsname
\fi
#2}}
\providecommand{\BIBdecl}{\relax}
\BIBdecl

\bibitem{roy2002learning}
D.~Roy and A.~Pentland, ``Learning words from sights and sounds: a
  computational model,'' \emph{Cognitive Science}, vol.~26, pp. 113--146, 2002.

\bibitem{Yu_Ballard_2004}
C.~Yu and D.~Ballard, ``A multimodal learning interface for grounding spoken
  language in sensory perceptions,'' \emph{ACM Transactions on Applied
  Perceptions}, vol.~1, pp. 57--80, 2004.

\bibitem{tenbosch_2009}
L.~ten Bosch, H.~Van~hamme, L.~Boves, and R.~K. Moore, ``A computational model
  of language acquisition: the emergence of words,'' \emph{Fundamenta
  Informaticae}, vol.~90, pp. 229--249, 2008.

\bibitem{Driesen_2011}
J.~Driesen and H.~Van~hamme, ``Modeling vocabulary acquisition, adaptation and
  generalization in infants using adaptive {B}ayesian {PLSA},''
  \emph{Neurocomputing}, vol.~74, pp. 1874--1882, 2011.

\bibitem{Rasanen_Rasilo_2015}
O.~Räsänen and H.~Rasilo, ``A joint model of word segmentation and meaning
  acquisition through cross-situational learning,'' \emph{Psychological
  Review}, vol. 122, pp. 792--829, 2015.

\bibitem{Mangin_2015}
O.~Mangin, D.~Filliat, L.~ten Bosch, and P.-Y. Oudeyer, ``{MCA-NMF}: Multimodal
  concept acquisition with non-negative matrix factorization,'' \emph{PLOS
  One}, 2015.

\bibitem{socher2014grounded}
R.~Socher, A.~Karpathy, Q.~V. Le, C.~D. Manning, and A.~Y. Ng, ``Grounded
  compositional semantics for finding and describing images with sentences,''
  \emph{{Transactions of the Association for Computational Linguistics}},
  vol.~2, pp. 207--218, 2014.

\bibitem{karpathy2015deep}
A.~Karpathy and F.~Li, ``Deep visual-semantic alignments for generating image
  descriptions,'' in \emph{{{IEEE} Conference on Computer Vision and Pattern
  Recognition ({CVPR} 2015)}}, June 7--12, 2015, Boston, MA, pp. 3128--3137.,
  2015, pp. 3128--3137.

\bibitem{harwath2015deep}
D.~Harwath and J.~Glass, ``Deep multimodal semantic embeddings for speech and
  images,'' in \emph{2015 IEEE Workshop on Automatic Speech Recognition and
  Understanding (ASRU)}.\hskip 1em plus 0.5em minus 0.4em\relax IEEE, 2015, pp.
  237--244.

\bibitem{harwath2016unsupervised}
D.~F. Harwath, A.~Torralba, and J.~R. Glass, ``Unsupervised learning of spoken
  language with visual context,'' in \emph{{Advances in Neural Information
  Processing Systems 29: Annual Conference on Neural Information Processing
  Systems (NIPS 2016)}}, December 5--10, 2016, Barcelona, Spain, pp.
  1858--1866., 2016, pp. 1858--1866.

\bibitem{synnaeve2014learning}
G.~Synnaeve, M.~Versteegh, and E.~Dupoux, ``Learning words from images and
  speech,'' in \emph{{28th Conference on Neural Information Processing Systems
  (NIPS) Workshop on Learning Semantics}}, December 8--13, 2014, Montreal,
  Canada., 2014.

\bibitem{chrupala2017representations}
G.~Chrupa{\l}a, L.~Gelderloos, and A.~Alishahi, ``Representations of language
  in a model of visually grounded speech signal,'' in \emph{{Proceedings of the
  55th Annual Meeting of the Association for Computational Linguistics (Volume
  1: Long Papers)}}, July 30--August 4, 2017, Vancover, Canada, pp. 613--622,
  2017, pp. 613--622.

\bibitem{kamper2019semantic}
H.~Kamper, G.~Shakhnarovich, and K.~Livescu, ``Semantic speech retrieval with a
  visually grounded model of untranscribed speech,'' \emph{{IEEE}/{ACM}
  Transactions on Audio, Speech and Language Processing}, vol.~27, pp. 89--98,
  2019.

\bibitem{alishahi2017encoding}
A.~Alishahi, M.~Barking, and G.~Chrupa{\l}a, ``Encoding of phonology in a
  recurrent neural model of grounded speech,'' in \emph{{Proceedings of the
  21st Conference on Computational Natural Language Learning ({C}o{NLL}
  2017)}}, August 3--4, 2017, Vancouver, Canada, pp. 368--378, 2017, pp.
  368--378.

\bibitem{harwath2019learning}
D.~Harwath, W.-N. Hsu, and J.~Glass, ``Learning hierarchical discrete
  linguistic units from visually-grounded speech,'' \emph{arXiv preprint
  arXiv:1911.09602}, 2019.

\bibitem{havard2019models}
W.~N. Havard, J.~Chevrot, and L.~Besacier, ``Models of visually grounded speech
  signal pay attention to nouns: {A} bilingual experiment on english and
  japanese,'' in \emph{{{IEEE} International Conference on Acoustics, Speech
  and Signal Processing ({ICASSP} 2019)}}, May 12--17, 2019, Brighton, UK, pp.
  8618--8622., 2019, pp. 8618--8622.

\bibitem{havard2019word}
------, ``Word recognition, competition, and activation in a model of visually
  grounded speech,'' in \emph{{Proceedings of the 23rd Conference on
  Computational Natural Language Learning (CoNLL 2019)}}, November 3--4, 2019,
  Hong Kong, China, pp. 339--348., 2019, pp. 339--348.

\bibitem{Chung2019a}
Y.~A. Chung, W.~N. Hsu, H.~Tang, and J.~Glass, ``{An unsupervised
  autoregressive model for speech representation learning},'' \emph{Proceedings
  of the Annual Conference of the International Speech Communication
  Association, INTERSPEECH}, pp. 146--150, 2019.

\bibitem{oord2018cpc}
\BIBentryALTinterwordspacing
A.~van~den Oord, Y.~Li, and O.~Vinyals, ``Representation learning with
  contrastive predictive coding,'' \emph{CoRR}, vol. abs/1807.03748, 2018.
  [Online]. Available: \url{http://arxiv.org/abs/1807.03748}
\BIBentrySTDinterwordspacing

\bibitem{chrupala2021visually}
G.~Chrupała, ``Visually grounded models of spoken language: A survey of
  datasets, architectures and evaluation techniques,'' 2021,
  \url{https://arxiv.org/abs/2104.13225}.

\bibitem{versteegh2015zero}
M.~Versteegh, X.~Anguera, A.~Jansen, and E.~Dupoux, ``The zero resource speech
  challenge 2015: Proposed approaches and results,'' \emph{Procedia Computer
  Science}, vol.~81, pp. 67--72, 12 2016.

\bibitem{dunbar2017zero}
E.~Dunbar, X.~N. Cao, J.~Benjumea, J.~Karadayi, M.~Bernard, L.~Besacier,
  X.~Anguera, and E.~Dupoux, ``The zero resource speech challenge 2017,'' 2017.

\bibitem{dunbar2019zero}
E.~Dunbar, R.~Algayres, J.~Karadayi, M.~Bernard, J.~Benjumea, X.-N. Cao,
  L.~Miskic, C.~Dugrain, L.~Ondel, A.~W. Black, L.~Besacier, S.~Sakti, and
  E.~Dupoux, ``The zero resource speech challenge 2019: Tts without t,'' 2019.

\bibitem{nguyen2020zero}
T.~A. Nguyen, M.~de~Seyssel, P.~Roz{\'e}, M.~Rivi{\`e}re, E.~Kharitonov,
  A.~Baevski, E.~Dunbar, and E.~Dupoux, ``The zero resource speech benchmark
  2021: Metrics and baselines for unsupervised spoken language modeling,''
  \emph{arXiv preprint arXiv:2011.11588}, 2020.

\bibitem{hsu_text-free_2020}
W.-N. Hsu, D.~Harwath, C.~Song, and J.~Glass, ``Text-{Free} {Image}-to-{Speech}
  {Synthesis} {Using} {Learned} {Segmental} {Units},'' in \emph{34th
  {Conference} on {Neural} {Information} {Processing} {Systems} ({NeurIPS})
  {Workshop} on {Self}-{Supervised} {Learning} for {Speech} and {Audio}
  {Processing}}, Dec. 2020.

\bibitem{Panayotov2015librispeech}
V.~{Panayotov}, G.~{Chen}, D.~{Povey}, and S.~{Khudanpur}, ``Librispeech: An
  asr corpus based on public domain audio books,'' in \emph{2015 IEEE
  International Conference on Acoustics, Speech and Signal Processing
  (ICASSP)}, 2015, pp. 5206--5210.

\bibitem{he_deep_2016}
K.~He, X.~Zhang, S.~Ren, and J.~Sun, ``Deep {Residual} {Learning} for {Image}
  {Recognition},'' in \emph{Proceedings of the {IEEE} conference on computer
  vision and pattern recognition}.\hskip 1em plus 0.5em minus 0.4em\relax
  Seattle, WA, USA: IEEE, Jun. 2016, pp. 770--778.

\bibitem{deng_imagenet_2009}
J.~Deng, W.~Dong, R.~Socher, L.-J. Li, K.~Li, and L.~Fei-Fei, ``{ImageNet}: {A}
  large-scale hierarchical image database,'' in \emph{2009 {IEEE} {Conference}
  on {Computer} {Vision} and {Pattern} {Recognition}}, Jun. 2009, pp. 248--255,
  iSSN: 1063-6919.

\bibitem{merkx_language_2019}
D.~Merkx, S.~L. Frank, and M.~Ernestus, ``Language {Learning} {Using} {Speech}
  to {Image} {Retrieval},'' in \emph{Proc. {Interspeech} 2019}, 2019, pp.
  1841--1845.

\bibitem{chrupala_symbolic_2019}
G.~Chrupa{\l }a, ``Symbolic {Inductive} {Bias} for {Visually} {Grounded}
  {Learning} of {Spoken} {Language},'' in \emph{Proceedings of the 57th
  {Annual} {Meeting} of the {Association} for {Computational}
  {Linguistics}}.\hskip 1em plus 0.5em minus 0.4em\relax Florence, Italy:
  Association for Computational Linguistics, Jul. 2019, pp. 6452--6462.

\bibitem{higy_textual_2020}
B.~Higy, D.~Elliott, and G.~Chrupa{\l }a, ``Textual {Supervision} for
  {Visually} {Grounded} {Spoken} {Language} {Understanding},'' in
  \emph{Findings of the {Association} for {Computational} {Linguistics}:
  {EMNLP} 2020}.\hskip 1em plus 0.5em minus 0.4em\relax Online: Association for
  Computational Linguistics, Nov. 2020, pp. 2698--2709.

\bibitem{cho_learning_2014}
K.~Cho, B.~van Merrienboer, C.~Gulcehre, D.~Bahdanau, F.~Bougares, H.~Schwenk,
  and Y.~Bengio, ``Learning {Phrase} {Representations} using {RNN}
  {Encoder}{\textendash}{Decoder} for {Statistical} {Machine} {Translation},''
  in \emph{Proceedings of the 2014 {Conference} on {Empirical} {Methods} in
  {Natural} {Language} {Processing} ({EMNLP})}.\hskip 1em plus 0.5em minus
  0.4em\relax Doha, Qatar: Association for Computational Linguistics, 2014, pp.
  1724--1734.

\bibitem{vandam_homebank_2015}
\BIBentryALTinterwordspacing
M.~VanDam, ``{HomeBank} {VanDam} {Public} 5-minute {Corpus},'' 2015, type:
  dataset. [Online]. Available:
  \url{http://homebank.talkbank.org/access/Public/VanDam-5minute.html}
\BIBentrySTDinterwordspacing

\bibitem{vandam_homebank_2015-1}
\BIBentryALTinterwordspacing
------, ``{HomeBank} {VanDam} {Public} {Daylong} {Corpus},'' 2015, type:
  dataset. [Online]. Available:
  \url{http://homebank.talkbank.org/access/Public/VanDam-Daylong.html}
\BIBentrySTDinterwordspacing

\bibitem{zhou2018towards}
L.~Zhou, C.~Xu, and J.~Corso, ``Towards automatic learning of procedures from
  web instructional videos,'' in \emph{Proceedings of the AAAI Conference on
  Artificial Intelligence}, vol.~32, no.~1, 2018.

\bibitem{sanabria18how2}
\BIBentryALTinterwordspacing
R.~Sanabria, O.~Caglayan, S.~Palaskar, D.~Elliott, L.~Barrault, L.~Specia, and
  F.~Metze, ``{How2:} a large-scale dataset for multimodal language
  understanding,'' in \emph{Proceedings of the Workshop on Visually Grounded
  Interaction and Language (ViGIL)}.\hskip 1em plus 0.5em minus 0.4em\relax
  NeurIPS, 2018. [Online]. Available: \url{http://arxiv.org/abs/1811.00347}
\BIBentrySTDinterwordspacing

\bibitem{miech19howto100m}
A.~Miech, D.~Zhukov, J.-B. Alayrac, M.~Tapaswi, I.~Laptev, and J.~Sivic,
  ``How{T}o100{M}: {L}earning a {T}ext-{V}ideo {E}mbedding by {W}atching
  {H}undred {M}illion {N}arrated {V}ideo {C}lips,'' in \emph{ICCV}, 2019.

\bibitem{AVA}
\BIBentryALTinterwordspacing
S.~Chaudhuri, J.~Roth, D.~Ellis, A.~C. Gallagher, L.~Kaver, R.~Marvin,
  C.~Pantofaru, N.~C. Reale, L.~G. Reid, K.~Wilson, and Z.~Xi, ``Ava-speech: A
  densely labeled dataset of speech activity in movies,'' in \emph{Proceedings
  of Interspeech, 2018}, 2018. [Online]. Available:
  \url{https://arxiv.org/pdf/1808.00606}
\BIBentrySTDinterwordspacing

\end{thebibliography}

\end{appendices}

\end{document}